%% file: neurips_2024.tex
\documentclass{article}

\usepackage[preprint]{neurips_2024}

\usepackage[utf8]{inputenc} 
\usepackage[T1]{fontenc}    
\usepackage{url}            
\usepackage{booktabs}       
\usepackage{amsfonts}       
\usepackage{nicefrac}       
\usepackage{microtype}      
\usepackage{xcolor}         

\usepackage{adjustbox}
\usepackage{colortbl}
\usepackage{fancyhdr}
\usepackage{silence}
\usepackage{numprint}
\usepackage{soul}
\usepackage{algorithm}
\usepackage{algpseudocode}
\usepackage{caption}
\usepackage{subcaption}
\usepackage{graphicx}
\usepackage{arydshln}
\usepackage{tabularray}
\usepackage{diagbox}
\usepackage{letltxmacro}
\usepackage{bbding}
\usepackage{makecell}

\usepackage{amsmath}
\usepackage{graphicx}

\title{EMMA: Your Text-to-Image Diffusion Model Can Secretly Accept Multi-Modal Prompts
}

\author{\textbf{Yucheng Han$^{1,2}$\thanks{Equal contributions. Work was done when Yucheng Han was a Research Intern at Tencent.} 
\quad
Rui Wang$^{2\ast}$\thanks{Project Leader.}
\quad 
Chi Zhang$^{2\ast}$ \thanks{Corresponding Author. }
\quad Juntao Hu$^{2}$ } \\ \textbf{ Pei Cheng$^2$ \quad Bin Fu$^2$ \quad Hanwang Zhang$^1$ }\vspace{0.3em} \\
{\normalsize $^1$Nanyang Technological University} \quad
{\normalsize $^2$ Tencent} \\
{\normalsize $^1$ \{yucheng002,~hanwangzhang\}@ntu.edu.sg}  \\
{\normalsize $^2$ \{raywwang,~johnczhang,~jetthu,~brianfu\}@tencent.com} \\
\url{https://tencentqqgylab.github.io/EMMA}
}

\begin{document}

\maketitle

\begin{figure}[h]
    \centering
    \includegraphics[width=.95\textwidth]{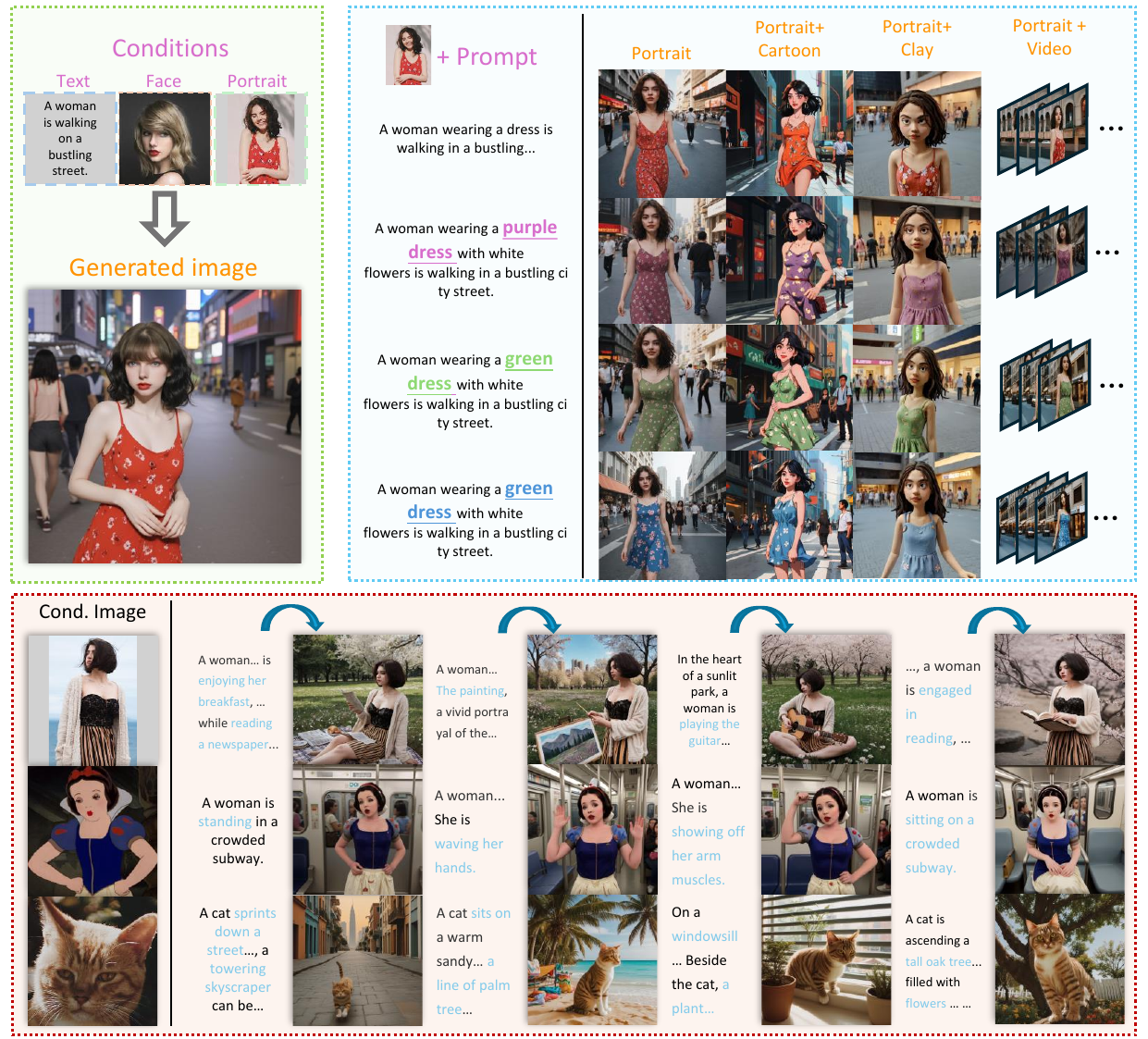}
    \caption{EMMA could compose multiple multi-modal conditions (on the top left branch) without further finetuning, while still maintaining strong text control over the generated results (bottom branch). Furthermore, EMMA could combine various existing diffusion models in communities without training.}
    \label{fig:teaser}
\end{figure}

\input{sections/0-abstract}
\input{sections/1-introduction}
\input{sections/2-related-work}
\input{sections/3-method}

\input{sections/4-experiments}

\input{sections/5-conclusion}

\clearpage
{
    \small
    \bibliographystyle{unsrtnat}
    \bibliography{citefile}
}

\appendix

\input{sections/x-supplymentary}


\end{document}

%% file: sections/0-abstract.tex
\begin{abstract}

Recent advancements in image generation have enabled the creation of high-quality images from text conditions. However, when facing multi-modal conditions, such as text combined with reference appearances, existing methods struggle to balance multiple conditions effectively, typically showing a preference for one modality over others. To address this challenge, we introduce EMMA, a novel image generation model accepting multi-modal prompts built upon the state-of-the-art text-to-image (T2I) diffusion model, ELLA.
EMMA seamlessly incorporates additional modalities alongside text to guide image generation through an innovative Multi-modal Feature Connector design, which effectively integrates textual and supplementary modal information using a special attention mechanism. 
By freezing all parameters in the original T2I diffusion model and only adjusting some additional layers, we reveal an interesting finding that the pre-trained T2I diffusion model can secretly accept multi-modal prompts.
This interesting property facilitates easy adaptation to different existing frameworks, making EMMA a flexible and effective tool for producing personalized and context-aware images and even videos.  
Additionally, we introduce a strategy to assemble learned EMMA modules to produce images conditioned on multiple modalities simultaneously, eliminating the need for additional training with mixed multi-modal prompts.
Extensive experiments demonstrate the effectiveness of EMMA in maintaining high fidelity and detail in generated images, showcasing its potential as a robust solution for advanced multi-modal conditional image generation tasks.

\end{abstract}

%% file: sections/1-introduction.tex
\section{Introduction}\label{sec:intro}

The field of image generation has recently experienced significant growth, driven by advancements from both academic and industrial researchers. Recent models, such as DALLE-3 and Stable Diffusion 3~\cite{sd3}, have elevated text-conditioned image generation to unprecedented levels. These models, requiring only simple textual instructions, demonstrate remarkable capability in generating high-quality images with intricate details. 
These approaches typically involve a classifier-free mechanism during the diffusion process to integrate conditions. For example, in the widely adopted Stable Diffusion, text prompts work as conditions of the diffusion network via cross-attention mechanisms to enable text-to-image translation.

Recent studies have also explored image generation conditioned on multi-modal prompts, which require simultaneous guidance from multiple modalities. For example, IP-Adapter~\cite{ye2023ip} guides image generation by referring to both image prompts and textual instructions, through developed cross-attention modules. Similarly, FaceStudio~\cite{yan2023facestudio} adopted a hybrid guidance framework and could utilize stylized images, facial images, and textual prompts as conditions for personalized portrait generation.
Based on these techniques, a variety of interesting applications have emerged, such as subject-driven image generation~\cite{pan2023kosmos, li2024blip, purushwalkam2024bootpig}, personalized image generation~\cite{wang2024instantid, yan2023facestudio}, and artistic portrait creation~\cite{ye2023ip}.
However, previous works employ distinct strategies for multi-modal prompts, and the genetic architecture of general multi-modal guided image generation remains unknown.

One of the main challenges facing current design paradigms is how to balance various conditions. During the image generation process, when multiple conditions are used, current methods may tend to favor certain conditions over others. For instance, it is observed that IP-Adapter~\cite{ye2023ip}, which relies on text prompts and image features as conditions, may predominantly be influenced by the image features. 
This can be attributed to the inherent limitations within the model architectures of existing methods, which do not effectively manage the varying complexities associated with different conditions. When training a model on multi-modal prompts, it often learns to control just one condition effectively, neglecting more challenging ones. This results in a bias towards easier conditions. For example, if a network is trained with both an object image and its description as conditions, it might overly rely on the image to generate object appearance, failing to adequately learn from the description. This issue highlights the need for strategies in training that ensure balanced learning across all conditions to maintain the model's versatility and fairness.
Furthermore, the scarcity of multi-modal training datasets in specialized domains exacerbates the issue. Taking subject-driven image generation as an example, a series of models (Kosmos-g~\cite{pan2023kosmos}, BootPig~\cite{purushwalkam2024bootpig}, SSR-Encoder~\cite{zhang2023ssr}) uses cropped object images to serve both as conditions and as the ground truth, which is a common practice in this area. However, models trained on such datasets are limited to a simple copy-paste functionality and may ignore the textual conditions. The absence of suitable training datasets becomes increasingly problematic with an increasing number of conditions. The limitations of model architecture and the lack of appropriate training datasets make it difficult to achieve a balanced approach for image generation models with multiple conditions.

To address the challenge above, we aim to design a more flexible paradigm for multi-modal guidance, that could well balance multiple conditions. 
In this paper, we introduce EMMA. Our proposed EMMA is built upon the state-of-the-art text-conditioned diffusion model ELLA~\cite{hu2024ella}, which trains a transformer-like module, named Perceiver Resampler, to connect text embeddings from pre-trained text encoders and pre-trained diffusion models for better text-guided image generation.
ELLA can effectively utilize pre-trained text and diffusion knowledge to achieve SOTA results in dense prompt-based image generation without the need to adjust their raw parameters. 
ELLA has strong text-to-image generation ability, and our proposed EMMA could merge information from other modalities into text features for guidance. This is inspired by Flamingo~\cite{alayrac2022flamingo}, a multi-modal large language model aiming at multi-modal understanding. Flamingo employs a strategy where it encodes images and text separately and integrates image features into text features using cross-attention within various transformer layers in the large language model. In this way, Flamingo adopts text as the primary carrier of information and integrates information from other modalities into LLM precisely for multi-modal understanding. 
Similarly, leveraging the transformer structure used by ELLA, which extracts features from the LLM to inject into SD, we introduce information from other modalities in the intermediate layers of these transformers to facilitate multimodal guidance.

In detail, to control the image generation process by modalities beyond text, EMMA incorporates our proposed Assemblable Gated Perceiver Resampler (AGPR), which leverages cross-attention to inject information from additional modalities beyond texts. In our design, the AGPR blocks are strategically interleaved with the blocks of the Perceiver Resampler of ELLA. This arrangement ensures an effective integration of multi-modal information. 
During training, we freeze the raw modules of ELLA to maintain the control ability of text conditions. Finally, we get a series of models based on different conditions, such as text features combined with facial features, and text features combined with object-level image features. 

Notably, EMMA is inherently designed to handle multi-modal prompts as conditions, allowing for the straightforward combination of different multi-modal configurations. This is achieved by the gate mechanism in our AGPR, which could control the way of injecting information from other modalities into the textual features. This advantage enables diverse and complex inputs to be synthesized into a unified generation framework without the need for additional training.
For example, image features can be utilized to depict the main subject, while finer-grained facial features provide identity information. 

As EMMA does not necessitate modifications to the underlying diffusion model, i.e. the U-net model or DiT~\cite{chen2023pixart, dit} model, 
it is readily compatible with a multitude of existing works based on the Stable Diffusion framework. By directly replacing the condition modules with EMMA, a series of interesting applications could be produced with no need for further training, such as Portrait generation, Cartoon generation, and subject-driven video generation shown in Figure~\ref{fig:teaser}. 

\begin{figure}[t]
    \centering
    \includegraphics[width=0.95\textwidth]{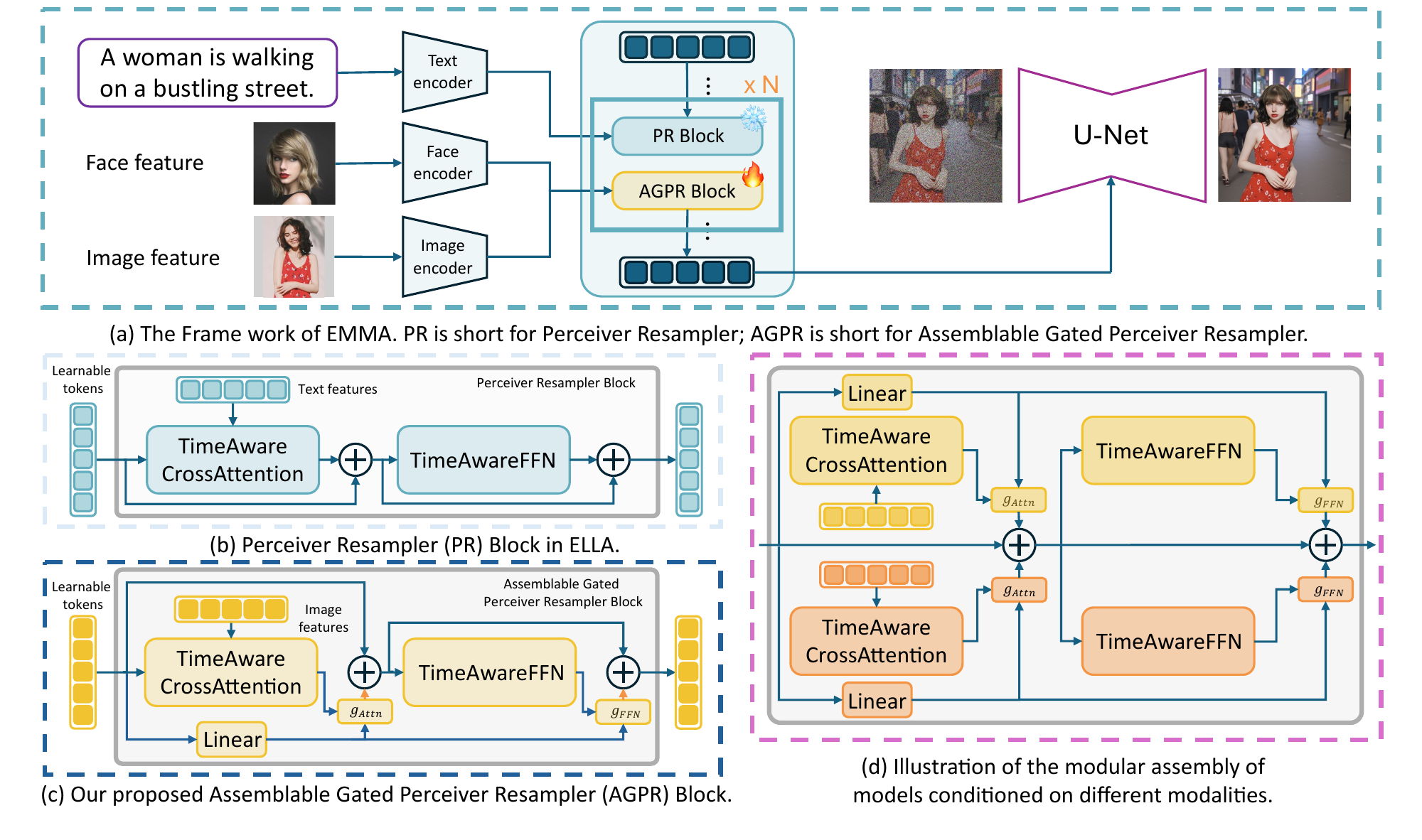}
    \caption{The model architecture of our proposed EMMA. (a) The framework of our EMMA. (b) The architecture of the Perceiver Resampler block proposed in ELLA~\cite{hu2024ella} (c) The architecture of our Assemblable Gated Perceiver Resampler block. The orange part is the novel part introduced in our AGPR block compared with the Perceiver Resampler block. (d) The pipeline of the composite process. }
    \label{fig:methods}
\end{figure}

Our key contributions are as follows:
\begin{enumerate}
    \item \textbf{Novel Integration Mechanism for Multi-modal prompts}: We introduce the EMMA, a pioneering approach that merges features of multi-modal prompts into the image generation process without compromising textual control. Our approach significantly enhances the flexibility and applicability of image generation by enabling the synergistic interaction of multiple modalities. This innovation allows for the creation of high-quality images that are responsive to a variety of input conditions.
    \item \textbf{Modular and Efficient Model Training}: Our framework facilitates the modular assembly of models conditioned on different modalities, streamlining the process and eliminating the need for retraining when new conditions are introduced. This efficient training procedure conserves resources and accelerates the model's adaptability to novel tasks.
    \item \textbf{Universal Compatibility and Adaptability}: EMMA works as a plug-and-play module without fine-tuning for a spectrum of existing and emerging models, including various image and video generation applications. Its compatibility with the Stable Diffusion framework and other models enhances its utility across diverse domains.
    \item \textbf{Robust Performance and Detail Preservation}: Through our experiments, we have confirmed the robustness of the EMMA model against various control signals, ensuring that it preserves both textual and visual details in the generated images. The model's architecture is designed to be scalable and flexible, accommodating a wide range of conditions and applications while maintaining high fidelity and quality.
\end{enumerate}

%% file: sections/2-related-work.tex
\section{Related Work} \label{sec:related_work}

\paragraph{Text-to-Image Diffusion Models.} Text-to-image diffusion models have made significant strides in producing high-quality and diverse images. These models depend on robust text encoders to interpret intricate image descriptions. Several models, such as GLIDE\cite{nichol2021glide}, LDM\cite{rombach2022high}, DALL-E 2\cite{ramesh2022hierarchical}, and Stable Diffusion\cite{rombach2022high,podell2023sdxl}, leverage the pre-trained CLIP\cite{radford2021learning} model to generate text embeddings. Other models like Imagen\cite{saharia2022photorealistic}, Pixart-$\alpha$\cite{chen2023pixart}, ELLA\cite{hu2024ella}, and DALL-E 3\cite{betker2023improving} employ large pre-trained language models, such as T5\cite{raffel2020exploring}, to enhance their understanding of text. Some models, including eDiff-I\cite{balaji2022ediffi} and EMU\cite{dai2023emu}, use a combination of both CLIP and T5 embeddings to improve their capabilities. ParaDiffusion\cite{wu2023paradiffusion} proposes fine-tuning the LLaMA-2\cite{touvron2023llama} model during diffusion model training and utilizing the fine-tuned language model text features as a condition. To further enhance the prompt following ability, we integrate large language models (LLM\cite{raffel2020exploring,touvron2023llama,zhang2024tinyllama}) with pre-trained CLIP-based models, using techniques such as TSC (Textual Style Control).

\paragraph{Subject-driven Image Generation.} 
This category includes studies focused on enhancing personalization and subject specificity in image generation through innovative techniques and architectures. Subject-Diffusion~\cite{ma2023subject} integrates text and image semantics for personalized generation without test-time fine-tuning. ELITE~\cite{wei2023elite} and FastComposer~\cite{xiao2023fastcomposer} reduce the need for fine-tuning by employing efficient encoding and attention mechanisms for personalized image generation.
BLIP-Diffusion~\cite{li2024blip} and Kosmos-G~\cite{pan2023kosmos} utilize pre-trained models for quick and effective personalized image generation. Unified Multi-Modal Latent Diffusion~\cite{ma2023unified} and IP-Adapter~\cite{ye2023ip} enhance image quality by integrating multimodal inputs to align images with textual descriptions. FaceStudio~\cite{yan2023facestudio}, InstantID~\cite{wang2024instantid}, and PhotoMaker~\cite{li2023photomaker} address the high resource demands of previous models and include features for identity preservation, critical for high-fidelity tasks like artistic portrait generation. 
The MoA (Mixture-of-Attention)~\cite{ostashev2024moa} uses a novel mechanism to separate subject and context for better image quality. 
BootPIG~\cite{purushwalkam2024bootpig} uses the reference net to introduce low-level information and achieves pixel-level control over generated images. 
The most recent and related work is SSR-Encoder~\cite{zhang2023ssr}, which uses cross-attention to inject image information into text features and supports selective feature extraction.

\paragraph{Optimization-based subject-driven image generation.} The paper \cite{gal2022image} introduces a method to personalize text-to-image generation through unique embeddings derived from user-provided images, enhancing the creation of unique concepts. Dreambooth \cite{ruiz2023dreambooth} describes a technique for fine-tuning text-to-image models to produce novel, contextualized images of a specific subject using a unique identifier. The paper \cite{liu2023cones} explores the concept of neurons in diffusion models that facilitate customized generation and efficient storage. A subsequent study \cite{liu2023cones} addresses synthesizing images with multiple subjects using text embeddings and spatial layouts to improve the quality and control of the synthesis.

%% file: sections/3-method.tex
\section{Methodology}\label{sec:method}

\subsection{Model Architecture}

The overall pipeline of EMMA is depicted in Figure~\ref{fig:methods} (a). 
Our model's conditions encompass two aspects. One is the textual feature, and the other is the customized image features, such as visual clip features or facial embeddings.\\
In EMMA, we inject text features through Perceiver Resampler blocks proposed by ELLA~\cite{hu2024ella} as shown in Figure~\ref{fig:methods} (b). 
The image features are perceived by our newly proposed module named Assemblable Gated Perceiver Resampler as shown in Figure~\ref{fig:methods} (c).

To be more specific, we categorize EMMA into three main components and describe them in detail. 

\textbf{Text Encoder:} T5~\cite{chung2024scaling} is equipped to understand rich textual content. Prior research has shown that T5 is adept at extracting textual features, which makes it well-suited for supplying textual features to downstream tasks.

\textbf{Image Generator:} In the realm of image generation, numerous researchers and practitioners have fine-tuned various models on a clip-specific basis, aligning with their specific goals and data types. We strive for our final network to ensure the generalization of features, thereby maximizing the use of the high-quality models prevalent in the community.

\textbf{Multi-modal Feature Connector:} The network architecture is depicted in Figure~\ref{fig:methods}. Drawing inspiration from Flamingo~\cite{alayrac2022flamingo} and ELLA, the connector consists of two alternating stacked network modules: the Perceiver Resampler and the Assemblable Gated Perceiver Resampler. 
The Perceiver Resampler is primarily tasked with integrating textual information, while the Assemblable Gated Perceiver Resampler is designed to incorporate additional information. These network modules use an attention mechanism to assimilate multimodal information into the learnable token embeddings, which are then supplied to the U-net as conditions.
We give the definitions of these blocks as follows. The connector contains $K$ learnable tokens, denoted by $Latent$. Time embeddings, textual features, and additional conditions are represented by $t$, $T$, and $C$, respectively.

The Perceiver Resampler block can be divided into two parts:

\begin{equation}
    L = L + \mathtt{TimeAwareAttn}(L, T, t),
\end{equation}
\begin{equation}
    L = L + \mathtt{TimeAwareFFN}(L, t).
\end{equation}

Here, $\mathtt{TimeAwareAttn}$ and $\mathtt{TimeAwareFFN}$ are custom attention and feedforward neural network (FFN) modules that utilize AdaLN to integrate time embeddings into the inputs. The advantages of this approach have been demonstrated by ELLA.

The Assemblable Gated Perceiver Resampler is formulated similarly:

\begin{equation}
    L = L + AttnGate \cdot \mathtt{TimeAwareAttn}(L, C, t),
\end{equation}
\begin{equation}
    L = L + FFNGate \cdot \mathtt{TimeAwareFFN}(L, t).
\end{equation}

In these equations, $AttnGate$ and $FFNGate$ are two sets of gates that regulate the feature integration. Their definitions are as follows:

\begin{equation}
    AttnGate = \lambda \cdot \mathtt{Linear}(L) \cdot A
\end{equation}
\begin{equation}
    FFNGate = \lambda \cdot \mathtt{Linear}(L) \cdot F
\end{equation}

Here, $\lambda$ is the gate scale, a fixed hyperparameter, and $A$ and $F$ are global gates. $\mathtt{Linear}(L)$ are separable gates.

\begin{figure}[t]
    \centering
    \includegraphics[width=0.9\textwidth]{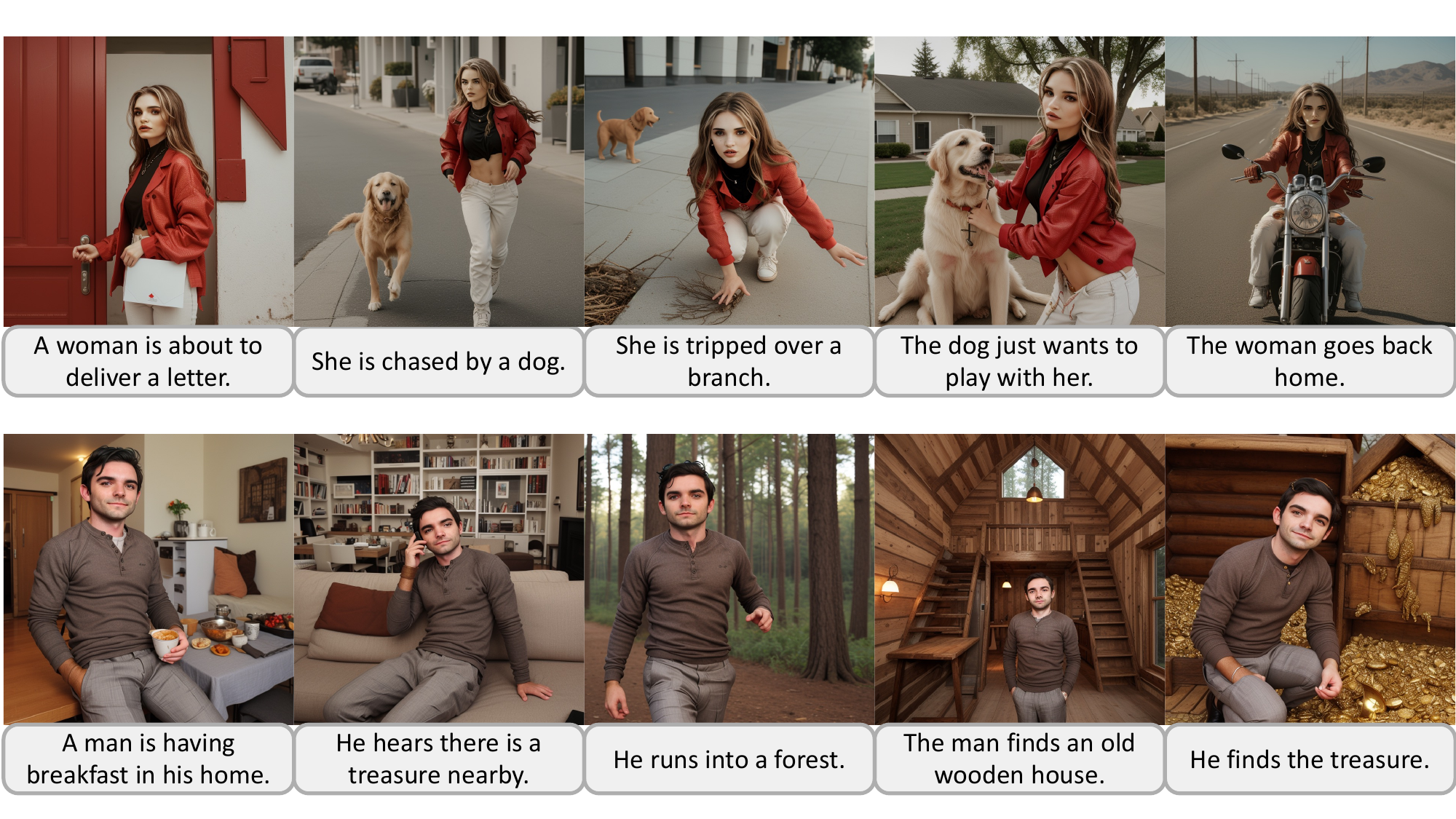}
    \caption{Images generated by our EMMA with portrait conditions. Two sets of images are generated for two separate stories. The first set of images is about a mailing woman chased by a dog. The second set of images is about a man finding treasures.}
    \label{fig:story_diffusion}
\end{figure}

\subsection{Image Generation with Multiple Conditions}

\textbf{Developing Text-to-Image Capability.} Through ELLA's training paradigm, we have developed a text-to-image model endowed with robust text-to-image capabilities. As illustrated in the first row of Figure~\ref{fig:final_visualization}, ELLA can generate images that strictly adhere to instructions, which forms the foundation for EMMA's multi-modal guidance.

\textbf{Selective Modular Feature Training.} To bolster the stability and enhance the final performance of the training process, we have integrated several innovative design elements into the network architecture. For example, the alternating structure between the Perceiver Resampler and the Assemblable Gated Perceiver Resampler is designed to limit the feature space of the network's intermediate layers. This prevents image information from imparting excessive prior knowledge that might compromise the text's control and disrupt the final generation outcomes. The Assemblable Gated Perceiver Resampler includes separated gates that enable the incorporation of additional features into a few trainable embeddings. 

\textbf{Assembling Modules for Multi-Condition Image Generation.} After establishing strong models for each individual condition, we have devised an innovative approach that enables the model to amalgamate existing modules and produce images conditioned by multiple factors. As depicted in the figure, we integrate the Assemblable Gated Perceiver Resampler. Without additional training, the model can synthesize all input conditions and generate novel outputs. This demonstrates the potential for image generation without relying on a pre-existing training dataset.

The process can be mathematically expressed as:
\begin{equation}
    L = L + \sum_{i} \lambda_i \cdot \mathtt{AttnGate}_i \cdot \mathtt{TimeAwareAttn}(L, C_i, t_i),
\end{equation}

\begin{equation}
    L = L + \sum_{i} \lambda_i \cdot \mathtt{FFNGate}_i \cdot \mathtt{TimeAwareFFN}(L, t_i).
\end{equation}

In this manner, various conditions can be applied to the image generation process without the need for further training.

%% file: sections/4-experiments.tex
\input{tables/style_comparison}
\input{tables/main_table}
\section{Experiments}\label{sec:exp}
\begin{figure}[t]
    \centering
    \includegraphics[width=0.9\textwidth]{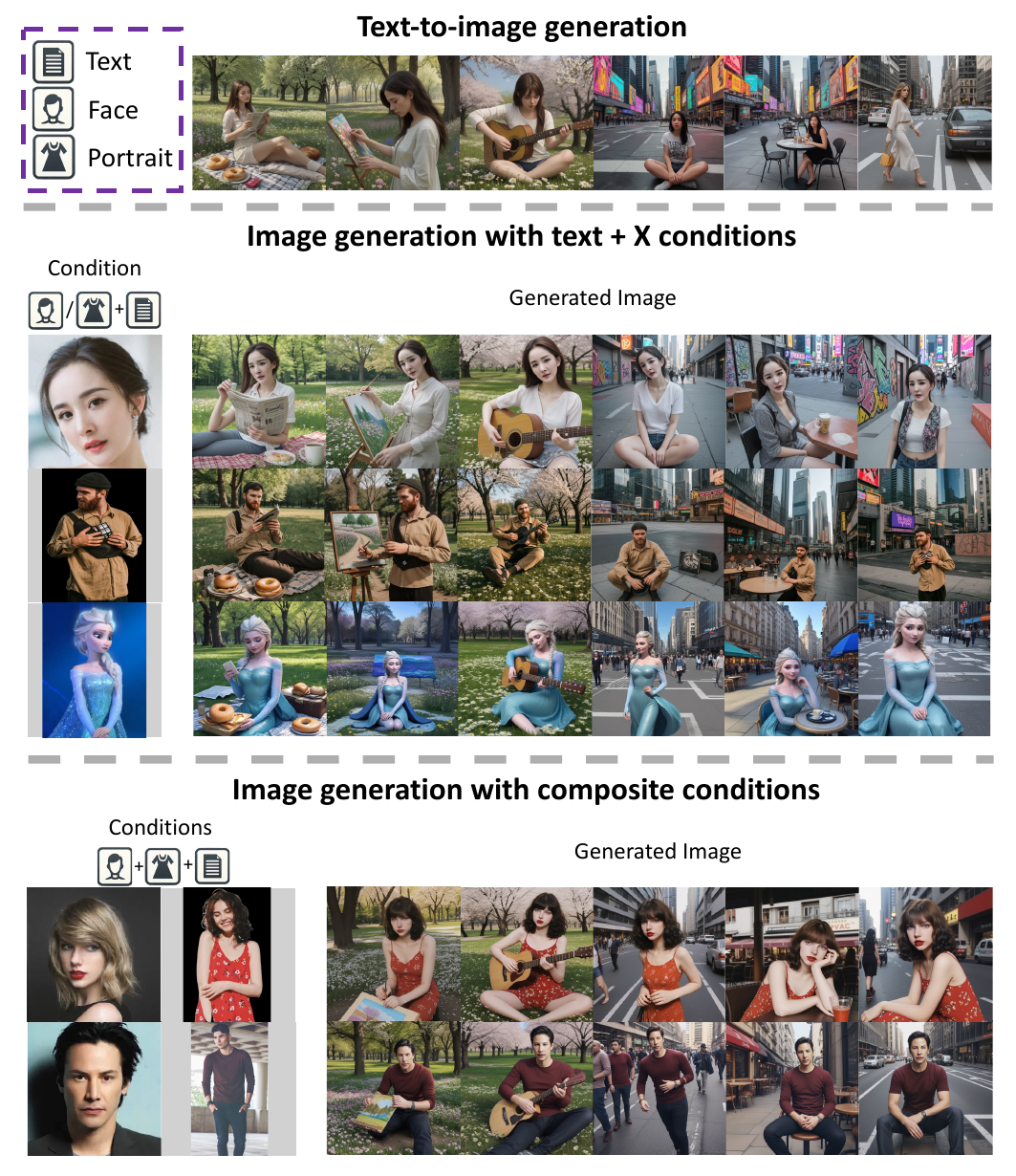}
    \caption{Visualization for our EMMA's generalization ability under different conditions. Each column shares the same text prompts. We show three kinds of conditions. The first row shows the results when there is only the text condition. The second row shows the results under multi-modal conditions, such as text plus face conditions and text plus portrait conditions. The bottom row shows the results under composite conditions.}
    \label{fig:final_visualization}
\end{figure}

\subsection{Dataset settings}
\textbf{Common object dataset.} We also collect datasets for common objects. Following ELLA~\cite{hu2024ella}, we filter images collected from LAION~\cite{schuhmann2022laion} and COYO~\cite{kakaobrain2022coyo-700m} with an aesthetic score over 6 and a minimum short edge resolution of 512 pixels. We generate several random masks to provide guidance for the central object. In this way, we can train the model on a large-scale dataset. 

\textbf{Portrait dataset.} We collect an internal dataset containing 400K images for 100K human IDs. Our EMMA targeted at portrait generation is fine-tuned on the internal dataset for 200K iterations. The test dataset uses 32 portraits and 20 prompts for each portrait, which are crawled from the Unsplash website and available under a use license. 

\subsection{Training Details}

We train our model based on the principles established by the Stable Diffusion 1.5, with modifications to suit our experimental requirements. The model employs a half-precision floating-point (fp16) data type for efficiency. We only change the conditioner and keep all the other key components unchanged, including the pre-trained Variational Autoencoder (VAE), the noise scheduler, and the UNet.

All the experiments are done on 8 A100 GPUs. We manage a total training batch size of 256, with micro batches of 16 per GPU. We implement gradient clipping at a value of 1.0.
The optimizer of choice is AdamW, which is configured with a learning rate of 0.0001. This setup includes betas of 0.9 and 0.999, an epsilon value of 1e-8, and a weight decay of 0.01. The learning rate is adjusted linearly from 10\% to 100\% over the course of 1000 iterations.
For different conditions, we employ different feature extractors and datasets, which are detailed in the Appendix. 

\subsection{Personalized Story Diffusion } Given specific character information, our proposed EMMA could generate different images according to the text instruction, which makes it possible to generate results telling a story while maintaining character consistency. As shown in Figure~\ref{fig:story_diffusion}, we can generate a series of images based on a given portrait following text instructions. The persons could do various actions, which benefit from the strong instruction-following abilities of EMMA.

\subsection{Quantitative Evaluation.}
\textbf{Style Conditioned Generation.} Following the evaluation settings of IP-Adapter~\cite{ye2023ip}, we evaluate the CLIP-T and CLIP-I scores of all methods on the COCO validation set. There are 5000 prompts in the validation set. We generate four images for each prompt as described in IP-Adapter~\cite{ye2023ip}.

\textbf{Portrait Generation.} We collect a dataset of portraits and construct 20 human action prompts based on the ActivityNet validation set. Building on this, we tested the generation capabilities of various subject-driven image generation methods and assessed the scores using the CLIP-T score and the DINO score metrics. Results are shown in Table~\ref{tab:main_table}, and our proposed EMMA achieves the highest score against previous methods. 

\textbf{Seperable Gate mechanism.} As shown in Table~\ref{tab:style_comparison}, we compare EMMA models trained under style conditions with and without separated gates. The EMMA with separated gates shows better performance, which is because such a design introduces finer control over different token embeddings. As observed in Figure~\ref{fig:gate_vis}, different tokens play different roles given specific conditions. Without the separated gates, the generated results will easily be influenced by unrelated token embeddings. 

\subsection{Visualization} 
\textbf{Different Conditions for Portrait Creation.} We have presented a variety of portrait generation outcomes. As seen in Figure \ref{fig:final_visualization}, our approach excels in maintaining key image elements like clothing and adheres closely to textual instructions. The top row illustrates the output of text-to-image generation, depicting a woman engaged in various activities across different settings. The middle row displays results from multi-modal image generation, where additional conditions such as facial or portrait traits yield images of a character that align with given instructions. The bottom row presents composite condition image generation, where we can produce images that follow instructions while retaining facial features from one image and portrait elements from another.

\begin{figure}[t]
    \centering
    \includegraphics[width=0.95\textwidth]{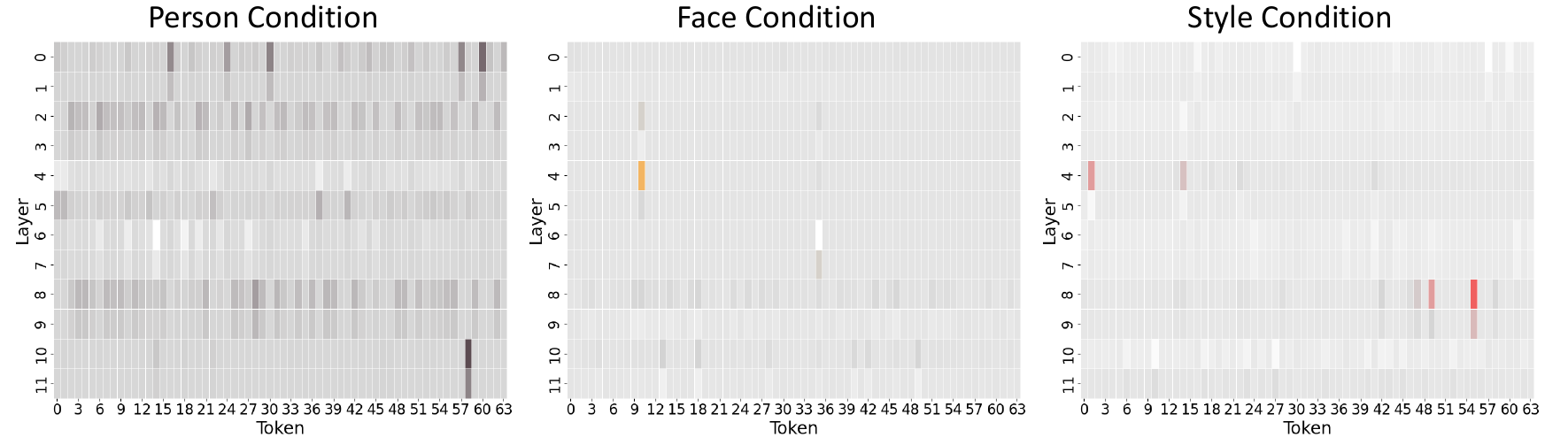}
    \caption{Visualization for gate values under different conditions. The horizontal axis is the token index, while the vertical axis is the depth of the Layer. We found that the gate values show sparsity features in different layers. We also found that models trained under different conditions pay attention to different tokens, which is the basis of module composition.}
    \label{fig:gate_vis}
\end{figure}

\textbf{Gate value visualization.} 
In our proposed EMMA, the gate design is a crucial module that enables free combination within our model. This design introduces an increased number of model parameters, enhancing the model's expressive capabilities. Furthermore, we observe a distinctive distribution of token indices of the significant gated values across various models. This unique pattern of token index distribution is crucial for the adaptability of our method, enabling flexible and unrestricted model integration. The visualization result is shown in Figure~\ref{fig:gate_vis}.

%% file: tables/style_comparison.tex
\begin{table}[tb]
    \centering
    \footnotesize
    \caption{Quantitative comparison for style conditioning of our proposed with other methods on the COCO validation set with four samples for every image. The best results are in \textbf{bold} (adapted from \cite{ye2023ip}).}
    \resizebox{0.85\linewidth}{!}{
    \begin{tabular}{lcccccc}
\toprule
  Style Method & \makecell[c]{Reusable to \\custom models} & \makecell[c]{Supports native \\control} & \makecell[c]{Multimodal \\prompts} & \makecell[c]{Composition \\ ability} & CLIP-T $\uparrow$ & CLIP-I $\uparrow$\\
\midrule
\emph{Training from scratch} \\
\midrule

Open unCLIP & \XSolidBrush & \XSolidBrush & \XSolidBrush & \XSolidBrush & \textbf{0.608} &\textbf{0.858} \\

Kandinsky-2-1 & \XSolidBrush & \XSolidBrush & \XSolidBrush & \XSolidBrush &0.599 &0.855 \\

Versatile Diffusion & \XSolidBrush & \XSolidBrush & \Checkmark & \XSolidBrush& 0.587 & 0.830\\
\midrule
\emph{ Fine-tuning from text-to-image model } \\
\midrule
SD Image Variations &\XSolidBrush & \XSolidBrush & \XSolidBrush &\XSolidBrush &0.548 &0.760 \\
SD unCLIP &\XSolidBrush & \XSolidBrush & \XSolidBrush & \XSolidBrush & \textbf{0.584} & \textbf{0.810} \\
\midrule
\emph{Adapters} \\
\midrule

Uni-ControlNet (Global Control) & \Checkmark & \Checkmark & \Checkmark &\XSolidBrush & 0.506 & 0.736 \\

T2I-Adapter (Style) & \Checkmark & \Checkmark & \Checkmark & \XSolidBrush & 0.485 & 0.648 \\

ControlNet Shuffle & \Checkmark & \Checkmark & \Checkmark & \XSolidBrush & 0.421 & 0.616 \\

IP-Adapter & \Checkmark & \XSolidBrush & \Checkmark & \XSolidBrush & 0.588 & 0.828 \\
\textbf{EMMA w/o separated gates} & \Checkmark & \Checkmark & \Checkmark & \Checkmark & 0.572 & 0.834  \\
\textbf{EMMA } & \Checkmark & \Checkmark & \Checkmark & \Checkmark & \textbf{0.594} & \textbf{0.860}  \\
\bottomrule
\end{tabular}}
\label{tab:style_comparison}
\end{table}

%% file: tables/main_table.tex
\begin{table}[t]
    \centering
    \footnotesize
    \caption{Quantitative comparison for portrait conditioned image generation. The best results are in \textbf{bold}.}
    \resizebox{0.7\linewidth}{!}{
    \begin{tabular}{ccccc}
    \toprule
        Method & IP-Adapter & BLIP-Diffusion & SSR-Encoder & \textbf{EMMA} \\\midrule
        CLIP-T $\uparrow$ & 49.54 & 56.27 & 58.75 & \textbf{64.00} \\
        DINO $\uparrow$ & 27.23 & 26.84 & 25.47 &  \textbf{29.86} \\ \bottomrule
    \end{tabular}
}
\label{tab:main_table}
\end{table}

%% file: sections/5-conclusion.tex
\section{Conclusion}\label{sec:conclusion}
In this paper, we propose EMMA, a multi-modal image generation model that has the potential to revolutionize the way images are created from diverse conditions. By integrating text and additional modalities through a unique Multi-modal Feature Connector, EMMA achieves a level of fidelity and detail in image generation that is unmatched by existing methods. Its modular allows for easy adaptation to various frameworks. Additionally, EMMA could composite existing modules to produce images conditioned on multiple modalities at the same time, eliminating the need for additional training. EMMA provides a highly efficient and adaptable solution for personalized image production.
In conclusion, EMMA's innovative approach to image generation sets a new benchmark for balancing multiple input modalities. As the field of generative models continues to evolve, EMMA is poised to become a cornerstone in the development of more sophisticated and user-friendly technologies, driving the next wave of innovation in AI-driven content creation.

\textbf{Limitations.} The current version of EMMA is only capable of processing English prompts. In the future, we will try to implement the same algorithm in diffusion models supporting multilingual prompts.

%% file: sections/x-supplymentary.tex
\newpage
\section{Appendix / supplemental material}

\subsection{Broader Impacts}\label{sec:broader_impacts}

The broader impacts of our novel multi-modal image generation model extend across various domains and societal aspects. Here, we provide a comprehensive reflection on the potential implications and ethical considerations associated with our advancements in conditional image generation.

\textbf{Impact on Creative Industries}: The ability to generate images from text and additional modalities can revolutionize various creative industries, from graphic design to film and gaming. While this may lead to concerns about job displacement, we anticipate that our model will primarily serve as a tool to augment the creative process, allowing professionals to achieve greater efficiency and explore new artistic frontiers.

\textbf{Accessibility and Empowerment}: By enabling the generation of high-fidelity images based on textual descriptions, our model can democratize the creation of visual content. This empowers individuals, including those without specialized artistic skills, to bring their ideas to life. We aim to make our technology accessible to a wide range of users, fostering creativity and innovation.

\textbf{Education and Research}: Our model can be a powerful educational tool, providing students and researchers with a means to visualize complex concepts and data. It can also facilitate scientific discovery by generating images that aid in the understanding of abstract or theoretical concepts, thereby enhancing learning and research outcomes.

\textbf{Ethical Use of Technology}: The potential for misuse of image generation technology, such as creating deepfakes or manipulating visual content for deceptive purposes, is a significant concern. We are dedicated to promoting the ethical use of our technology and are actively developing safeguards against such misuse. This includes:

\begin{itemize}
\item Watermarking and Traceability: Implementing features that allow the traceability of generated images, preventing unauthorized use and ensuring accountability.

\item Ethical Guidelines: Establishing clear guidelines for the ethical use of our model, emphasizing the importance of transparency and honesty in the generation and dissemination of images.

\item Collaboration with Stakeholders: Engaging with artists, content creators, and legal experts to develop a robust framework that protects intellectual property and ensures fair use.

\item Public Awareness: Educating the public about the capabilities and limitations of our technology, promoting responsible use and critical thinking regarding the authenticity of visual content.

\end{itemize}

\textbf{Environmental Considerations}: We are cognizant of the environmental impact associated with the computational requirements of AI models. Our approach to feature integration and the use of time embeddings aim to reduce the computational footprint, aligning with our commitment to sustainable AI development.

In conclusion, while our multi-modal image generation model presents exciting opportunities for innovation and creativity, it also comes with a set of ethical and societal responsibilities. We are dedicated to addressing these challenges proactively, ensuring that our technology is developed and used in a manner that is beneficial, responsible, and respectful of diverse societal values.

\subsection{Safeguards}\label{sec:safeguards}
\begin{enumerate}
    \item During training, our model utilized Stable Diffusion 1.5 which is capable of detecting NFSW content. This could prevent our model from generating and learning NFSW images. 
    \item The source of our internal dataset could guarantee that there is not any NFSW content. 
\end{enumerate}

\subsection{License}\label{sec:license}
\begin{enumerate}
    \item Stable Diffusion 1.5:  The CreativeML OpenRAIL M license.
    \item LAION: MIT License.
    \item COYO: CC-BY-4.0 License. 
    \item ELLA: Apache-2.0 license.
\end{enumerate}

\subsection{More visualization}

\begin{figure}[t]
    \centering
    \includegraphics[width=0.95\textwidth]{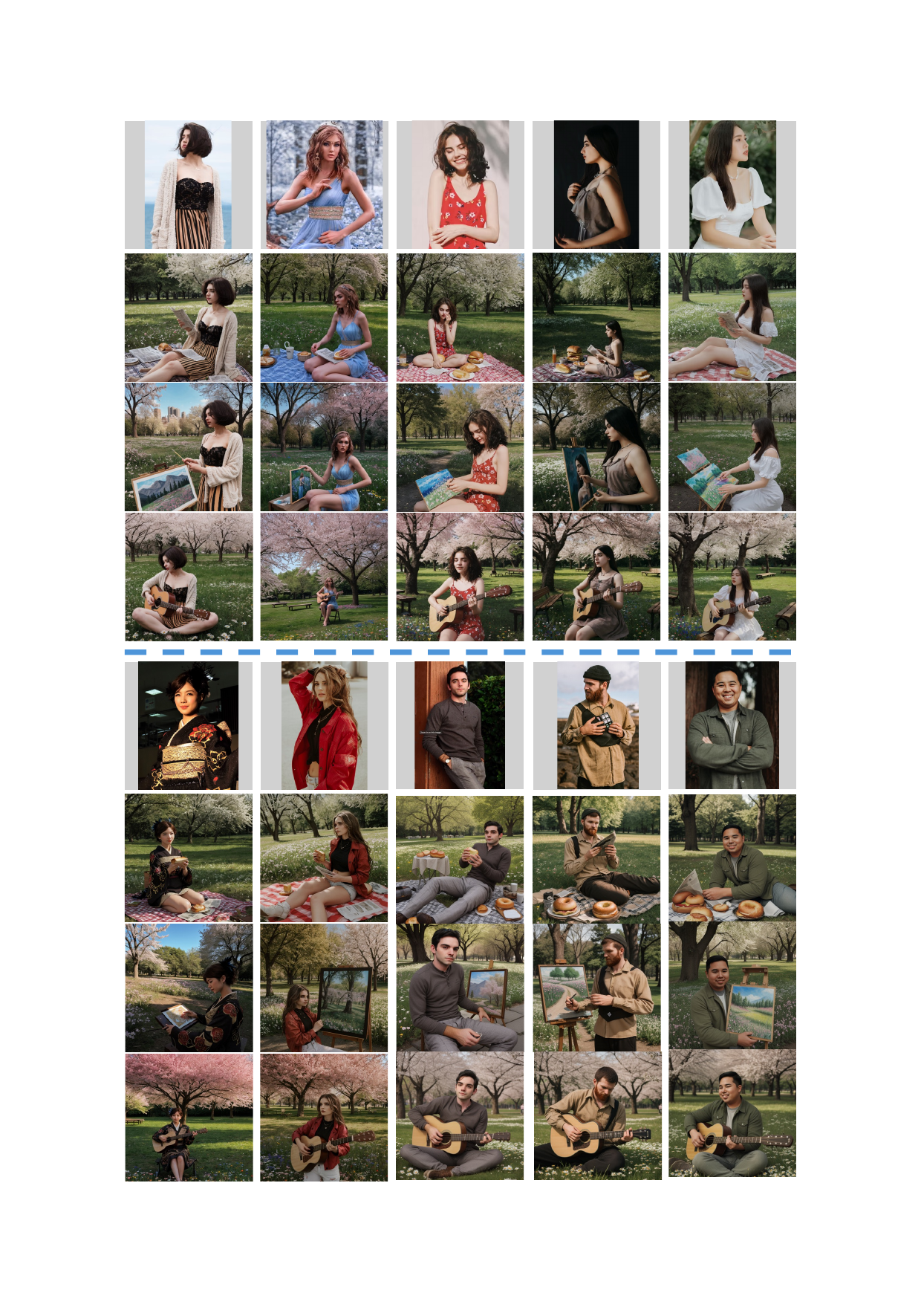}
    \caption{Illustration of more portrait generation examples. The first row is the condition image, and the following rows show the results of EMMA. The prompts for those examples are listed in our Appendix.}
    \label{fig:supplymentary}
\end{figure}

More visualization using portrait condition is shown in Figure~\ref{fig:supplymentary}. We show portrait generation results for both males and females. They all share the same generation prompts as those in Figure~\ref{fig:teaser}.

The prompts are listed below:
\begin{enumerate}
    \item A person sits on a checked picnic blanket in the lush, green park, surrounded by blooming wildflowers and tall trees. She is enjoying her breakfast, which consists of a toasted bagel with cream cheese and a steaming cup of coffee while reading a newspaper held delicately in her hand. The sun peeks through the branches, casting dappled shadows across the scene.
    \item a person is deeply engrossed in her artistic endeavor within a serene park surrounded by blossoming wildflowers and towering trees. The painting, a vivid portrayal of the park's essence, captures the interplay of light and shadow as the sun's rays dance through the foliage above. The tranquil setting enhances her focus, as the natural beauty of the park becomes an integral part of her creation.
    \item In the heart of a sunlit park, a person is playing guitar. Around her, vibrant pink cherry blossoms bloom profusely from their branches, creating a canopy of soft, delicate petals overhead. The lush green grass below is sprinkled with a tapestry of multi-colored wildflowers swaying gently in the breeze. A few nearby benches invite passersby to pause and enjoy the harmonious blend of nature and music. 
\end{enumerate}

\begin{figure}[t]
    \centering
    \includegraphics[width=0.9\textwidth]{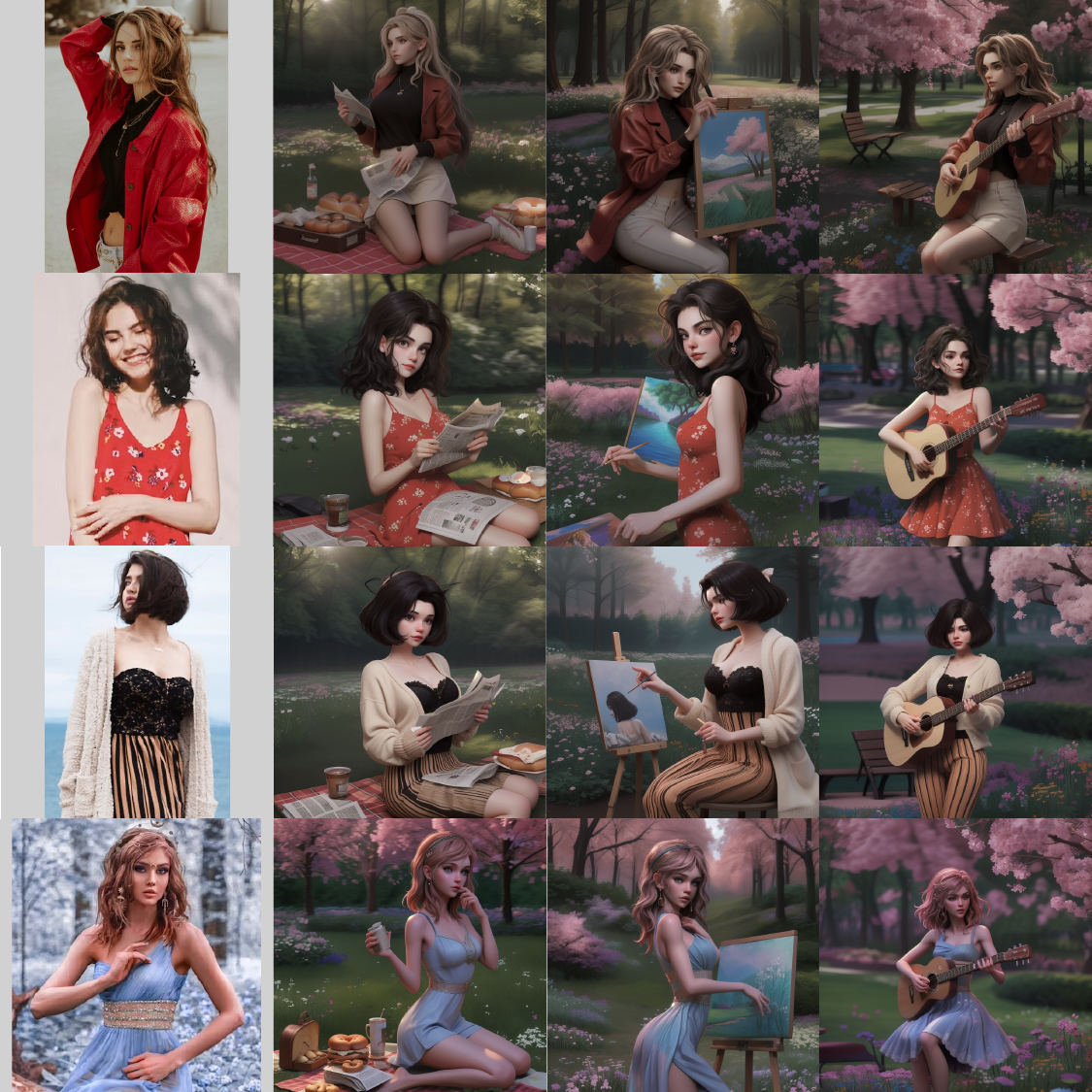}
    \caption{Generated results using EMMA and ToonYou's U-net. The first column is the condition image, and the following columns show the corresponding generated results. The prompts are also the same as Figure~\ref{fig:supplymentary}.}
    \label{fig:supplymentary_toonyou}
\end{figure}
\begin{figure}[t]
    \centering
    \includegraphics[width=0.9\textwidth]{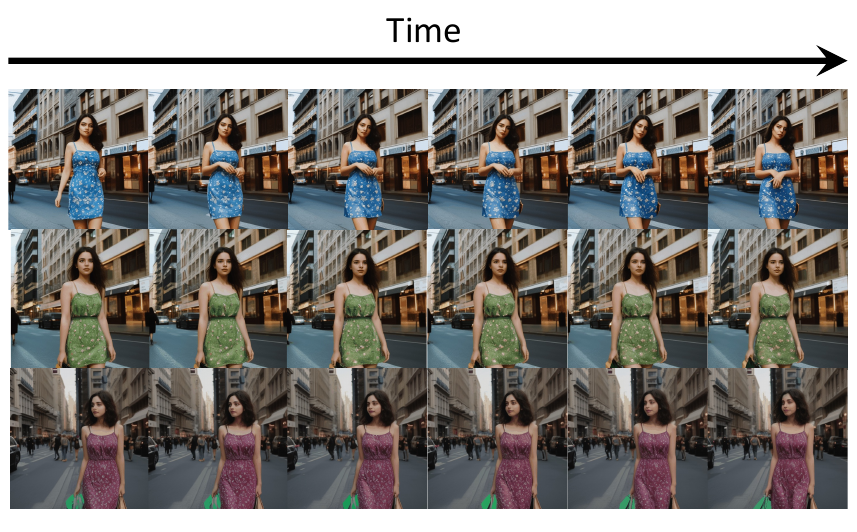}
    \caption{Generated results using EMMA and AnimateDiff's U-net. The conditional image and prompts are shown in Figure~\ref{fig:teaser}. }
    \label{fig:supplymentary_video}
\end{figure}

\subsection{Adaptation to existing extensions in community.} 
Since our proposed EMMA does not require training the diffusion models, we can utilize commonly used community-based diffusion models trained on CLIP text features, such as the picXreal and ToonYou models, which are representative of portrait and anime styles, respectively. Furthermore, our model can even be transferred to results from models like animatediff, which are secondary developments based on diffusion models. The results on these open-source communities are illustrated in Figure~\ref{fig:supplymentary_toonyou} and Figure~\ref{fig:supplymentary_video}.

\subsection{More Training Details}

\subsubsection{Training Settings for Different Conditions}
\paragraph{Text features plus common object features.} We train the model on our collected common object dataset for 200K iterations. The image feature extractor is CLIP-H/14, and we send both the global features and local features as the key and value features for cross-attention. The weights of this model also work as the initialization for the models conditioned on text features plus portrait features. 

\paragraph{Text features plus style features.} The model is trained on the common object dataset. The image features are also collected by CLIP-H/14 but only use the global features. The image features are then projected to 4 tokens by an extra linear layer. All the data processing procedures follow the IP-Adapter \cite{ye2023ip}.

\paragraph{Text features plus face features.} The model is trained on our own collected facial dataset for 200K iterations. We first detect and use only the face area for feature processing. Then we use AdaFace~\cite{kim2022adaface} for feature extraction and use them as the key and value features.

\subsubsection{More Ablations}

\textbf{Freeze Perceiver Resamplers.} Freezing the Perceiver Resamplers is an essential method for constructing effective multi-modal guidance. During training, we freeze the parameters of Perceiver Resamplers to keep the text following ability. Not freezing these layers will make it impossible for the composite of different EMMA models.

\textbf{Different assemble methods.} 
Our EMMA architecture enables the fusion of models from different conditions to form new models. Since these models do not require training, how to merge them becomes a question worth designing and contemplating. In addition to the combination methods outlined in our paper, such as those in formulas 3 and 4, we have designed several groups of results. Experimental results demonstrate that our method can significantly better integrate model characteristics. The way we merge models is also significantly related to the distinct patterns in the distribution of gate values.

\textbf{Object-centric mask.} During training and inference, we add an object-centric mask to avoid the influence of background image information.

\clearpage